%% file: newsroom.tex
\documentclass[11pt,a4paper]{article}
\usepackage[hyperref]{style/naaclhlt2018}
\usepackage{times, latexsym, url}

\aclfinalcopy
\def\aclpaperid{***}

\title{\DATASET: A Dataset of 1.3 Million Summaries \\ with Diverse Extractive Strategies}

\author{
Max Grusky$^{1,2}$, Mor Naaman$^2$, Yoav Artzi$^{1,2}$ \\
$^1$Department of Computer Science, $^2$Cornell Tech \\
Cornell University, New York, NY 10044 \\
{\tt \{grusky@cs, mor@jacobs, yoav@cs\}.cornell.edu }
}

\date{}

\input{sections/definitions}

\begin{document}

\setlength{\abovedisplayskip}{-0.5em}
\setlength{\belowdisplayskip}{0.5em}

\maketitle

\input{sections/0abstract}
\input{sections/1introduction}
\input{sections/2related}
\input{sections/3collection}
\input{sections/4analysis}
\input{sections/5methods}
\input{sections/6evaluation}
\input{sections/7conclusion}

\bibliographystyle{style/acl_natbib}
\bibliography{references}

\input{sections/appendix}

\end{document}

%% file: sections/definitions.tex
\usepackage{graphicx, xspace, todonotes, pbox}
\usepackage{amsmath, amssymb, amsthm, amssymb}
\usepackage{booktabs, tablefootnote, multirow, enumerate, enumitem}
\usepackage{algorithm, algpseudocode}

\usepackage[normalem]{ulem}
\usepackage[framemethod=tikz]{mdframed}

\usepackage{float}
\usepackage{stfloats}

\algtext*{EndFunction}
\algtext*{EndWhile}
\algtext*{EndIf}

\makeatletter
\g@addto@macro\small{
    \setlength\abovedisplayskip{-8pt}
    \setlength\belowdisplayskip{-8pt}
    \setlength\abovedisplayshortskip{-8pt}
    \setlength\belowdisplayshortskip{-4pt}
}
\makeatother

\usepackage{titlesec}
\titlespacing*{\paragraph} {0pt}{0.2ex plus 1ex minus .2ex}{1em}

\usepackage[skip=3pt]{caption}
\DeclareCaptionFormat{capfont}{\fontsize{10}{11}\selectfont#1#2#3}
\usepackage{subcaption}

\newcommand{\ul}[2]{{\color{#1}\uline{#2}}}
\newcommand{\ulb}[1]{\ul{black}{#1}}
\newcommand{\ulr}[1]{\ul{red}{#1}}

\newcommand{\summaryarticle}[2]{

	\vspace{-0.75em}

	\begin{center}
		\begin{tabular}{|p{\linewidth - 1em}|}
			\hline
			{\small \vspace{-0.25em} \textbf{Example Summary:} #1 \vspace{0.5em}}
			\\ \hline
			{\small \vspace{-0.25em} \textbf{Start of Article:} #2 [...] \vspace{0.5em}}
			\\ \hline
		\end{tabular}
	\end{center}

	\vspace{0.5em}

}

\newcommand{\footurl}[1]{\footnote{\url{#1}}}

\newcommand{\nb}[1]{\mbox{#1}}

\newcommand{\x}[1]{\textbf{#1}}
\newcommand{\y}[1]{\textbf{#1}}

\newcommand{\DATASET}{\textsc{Newsroom}\xspace}

\newcommand{\WEBSITE}{\href{http://lil.nlp.cornell.edu/newsroom/}{lil.nlp.cornell.edu/newsroom}\xspace}

\newcommand{\TextRank}{\nb{TextRank}\xspace}

\newcommand{\FIGUREEXAMPLESUMMARIES}{

	\begin{figure}
		\footnotesize
		\textbf{DUC} \vspace{-6pt}	
		\summaryarticle{
		\ulb{Floods hit} \emph{north} Mozambique \ulb{as aid} to flooded \emph{south continues}
	}{
		MAPUTO, Mozambique (AP) --- Just \ulb{as~aid} agencies were making headway in feeding hundreds of thousands displaced by flooding in southern and central Mozambique, new \ulb{floods hit} a remote northern region Monday. The Messalo River overflowed
	}\vspace{-5pt}	
		\textbf{Gigaword} \vspace{-6pt}	
		\summaryarticle{
		Seve \emph{gets} invite to \ulb{US Open}
	}{
		Seve Ballesteros will be playing in next month's \ulb{US Open} after all.
		The USGA decided Tuesday to give the Spanish star a special exemption.
		American Ben Crenshaw was also given a special exemption by the United States Golf Association.
		Earlier this week
	}\vspace{-5pt}	
		\textbf{New York Times Corpus}\vspace{-6pt}	
		\summaryarticle{
		\emph{Annual} \ulb{New York} City \ulb{Toy Fair} opens \ulb{in Manhattan}; \ulb{feud between} \ulb{Toy Manufacturers of America and its landlord at} \ulb{International Toy Center} \emph{leads} to \ulb{confusion and} \emph{turmoil} as registration \emph{begins}; \emph{dispute discussed}.
	}{
		There was toylock when the \ulb{Toy Fair} opened \ulb{in Manhattan} yesterday. The reason? A family \ulb{feud between} the \ulb{Toy Manufacturers of America and its landlord at} Fifth Avenue and 23d Street. Toy buyers and exhibitors arriving to attend the kickoff of the
	}\vspace{-5pt}	
		\textbf{CNN / Daily Mail}\vspace{-6pt}	
		\summaryarticle{
		\vspace{0.25em}
		\begin{itemize}[itemsep=0.05em, topsep=0.2em, leftmargin=1.5em]
			\item \ulb{Eight Al Jazeera journalists} are \ulb{named on} \emph{an} Egyptian \ulb{charge sheet}, the network \emph{says}
			\item \ulb{The eight} were \emph{among} \ulb{20 people} named
		`Most \ulb{are not employees of Al Jazeera},'' \ulb{the network said}
			\item \ulb{The eight} \ulb{include three} journalists jailed \ulb{in Egypt}
		\end{itemize}
		\vspace{-1.5em}
	}{
		Egyptian authorities have served Al Jazeera with a \ulb{charge sheet} that identifies eight of its staff on a list of \ulb{20 people} -- all believed to be journalists -- for allegedly conspiring with a terrorist group, \ulb{the network said} Wednesday. The 20 are wanted by Egyptian
	}\vspace{-5pt}	
	\caption{Example summaries for existing datasets.}
	\label{figure:datasets}
	\end{figure}

}

\newcommand{\SUBSETFIGURE}{

	\begin{figure}[ht]

	\centering
	\begin{tabular}{| p{\linewidth - 1.2em} |}
		\hline \\ [-0.9em]
		{\footnotesize \textbf{Abstractive Summary:}
		\ulr{South African} photographer \ulr{Anton Hammerl}, \emph{missing} in Libya \emph{since}~April~\emph{4th}, was \emph{killed} in Libya \emph{more than} a~\emph{month~ago}.}
		\\ [0.4em] \hline \\ [-0.9em]
		{\footnotesize \textbf{Mixed Summary:}
		A major climate protest \ulr{in New York} \ulr{on Sunday} \emph{could} mark \ulr{a seminal} \emph{shift} \ulr{in~the} \emph{politics} of \ulr{global warming}, just \emph{ahead} \ulr{of~the} \emph{U.N.} \ulr{Climate Summit}.}
		\\ [0.4em] \hline \\ [-0.9em]
		{\footnotesize \textbf{Extractive Summary:}
		\ulr{A person familiar with the search} \emph{tells} \ulr{The Associated Press} that \ulr{Texas has} \emph{offered} \ulr{its head coaching job to Louisvilles Charlie Strong and he is expected to accept}.}
		\\ [0.4em] \hline
	\end{tabular}

	\caption{
		\DATASET summaries showing different extraction strategies,
		from time.com, mashable.com, and foxsports.com.
		Multi-word phrases shared between article and summary are underlined.
		Novel words used only in the summary are italicized.
	}

	\label{figure:examples}

	\end{figure}

}

\newcommand{\OVERVIEWTABLE}{
  \begin{table}[t]

    \vspace{-1em}

    \centering
    \small

    \begin{tabular}{l r}
      \\ \midrule
      Dataset Size             & 1,321,995 articles
      \\
      Training Set Size        & 995,041 articles
      \\ \midrule
      Mean Article Length   & 658.6 words
      \\
      Mean Summary Length   & 26.7 words
      \\ \midrule
      Total Vocabulary Size    & 6,925,712 words
      \\
      Occurring 10+ Times       & 784,884 words
      \\ \midrule
    \end{tabular}

    \caption{Dataset Statistics}
    \label{table:statistics}

    \vspace{-0.5em}

  \end{table}
}

\newcommand{\FRAGMENTSALGORITHM}{
	\begin{figure}

		\begin{mdframed}[
		    userdefinedwidth=19em,
		    innerleftmargin=-0.5em,
		]
		
		\begin{algorithmic}[0]

            \small

	        \Function{$\mathcal{F}$}{$A, S$}

				\State $\mathcal{F} \gets \emptyset, ~ \langle i, j \rangle \gets \langle 1, 1 \rangle$

				\While{$i \le |S|$}

					\State $f \gets \langle ~ \rangle$ 

					\While{$j \le |A|$}

						\If{$s_i = a_j$}

							\State $\langle i', j' \rangle \gets \langle i, j \rangle$

							\While{$s_{i'} = a_{j'}$}

								\State $\langle i', j' \rangle \gets \langle i' + 1, j' + 1 \rangle$

							\EndWhile

							\If{$|f| < (i' - i - 1)$}

								\State $f \gets \langle s_i ~ \cdots ~ s_{i'-1} \rangle$

							\EndIf

							\State $j \gets j'$

						\Else

							\State $j \gets j + 1$

						\EndIf

					\EndWhile

					\State $\langle i, j \rangle \gets \langle i + \max\{|f|, ~ 1\}, ~ 1 \rangle$
					\State $\mathcal{F} \gets \mathcal{F} \cup \{ f \}$

				\EndWhile

				\State \textbf{return} $\mathcal{F}$

	        \EndFunction
	    \end{algorithmic}

	    \end{mdframed}

	    \caption{
		    Procedure to compute the set $\mathcal{F}(A, S)$ of extractive phrases in summary $S$ extracted from article $A$.
		    For each sequential token of the summary, $s_i$, the procedure iterates through tokens of the text, $a_j$.
		    If tokens $s_i$ and $a_j$ match, the longest shared token sequence after $s_i$ and $a_j$ is marked as the extraction starting at $s_i$.
	    }

	    \label{figure:fragments}

	\end{figure}
}

\newcommand{\HISTOGRAMFIGURE}{

	\begin{figure*}

        \centering

		\includegraphics[width=\textwidth - 8pt]{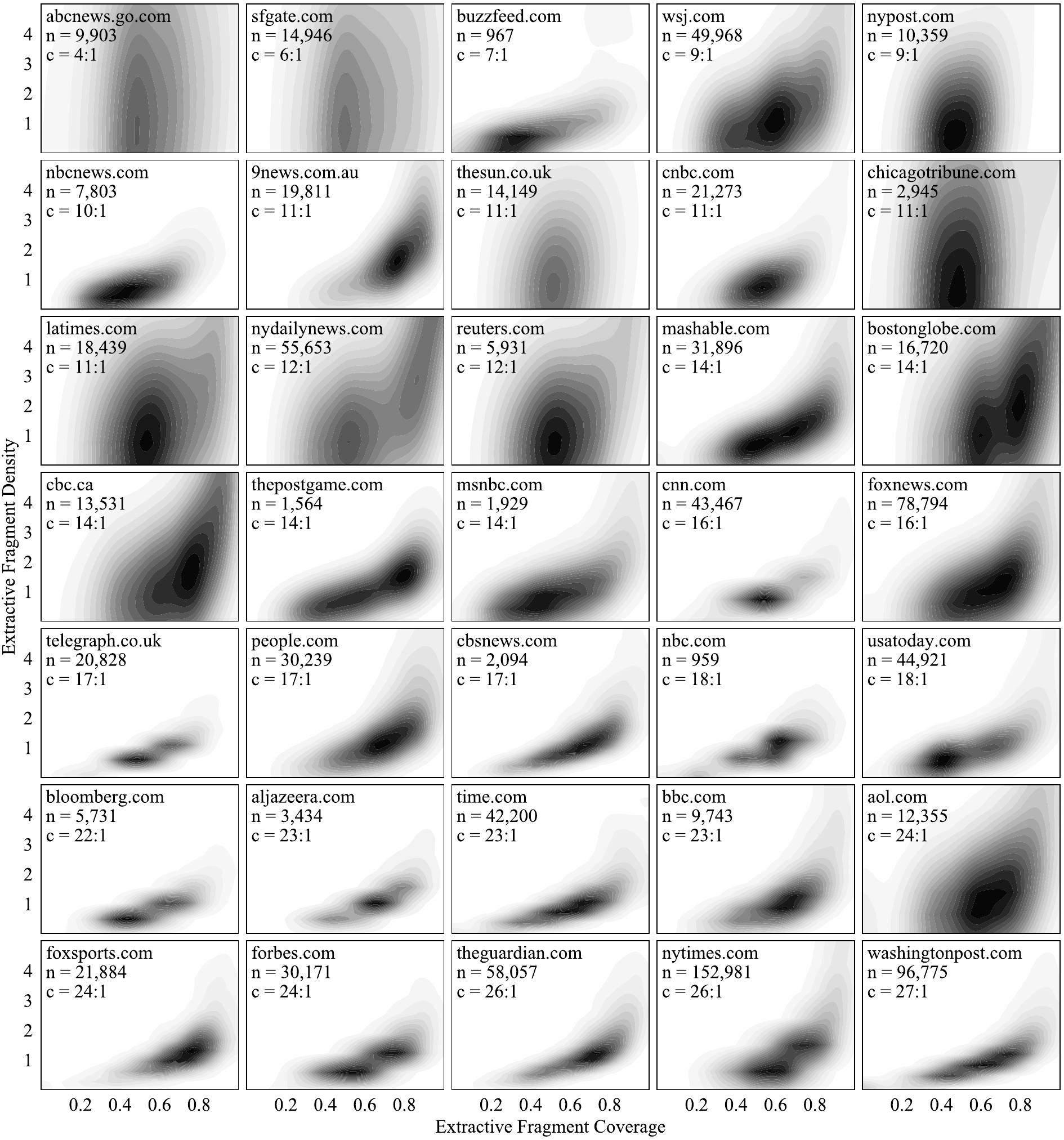}
		\includegraphics[width=\textwidth - 8pt]{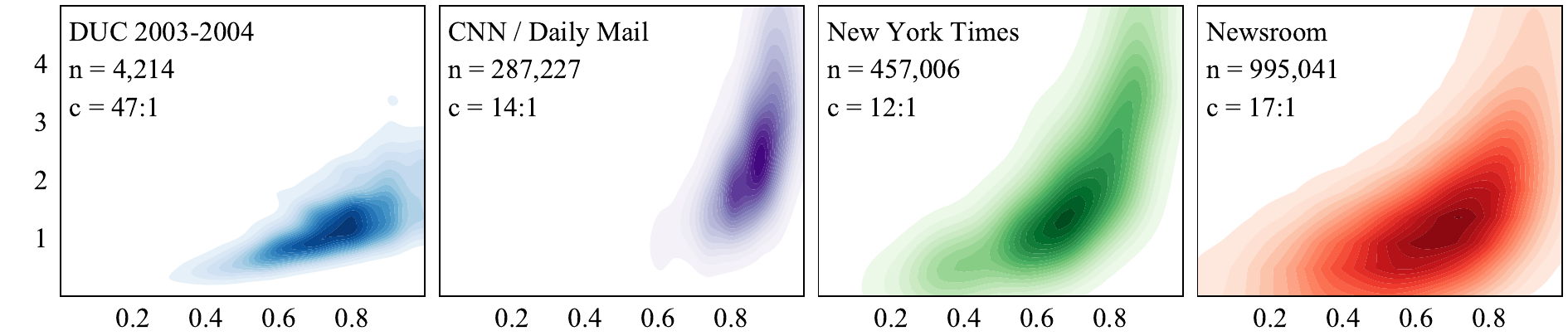}
		\includegraphics[width=\textwidth - 8pt]{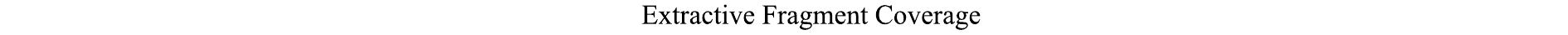}

		\caption{
			Density and coverage distributions across the different domains and existing datasets. 
			\DATASET contains diverse summaries that exhibit a variety of summarization strategies.
			Each box is a normalized bivariate density plot of extractive fragment coverage (x-axis) and density (y-axis), the two measures of extraction described in Section~\ref{section:analysis:dimensions}.
			The top left corner of each plot shows the number of training set articles $n$ and the median compression ratio $c$ of the articles.
			For DUC and New York Times, which have no standard data splits, $n$ is the total number of articles.
			\emph{Above, top left to bottom right:} Plots for each publication in the \DATASET dataset. We omit TMZ, Economist, and ABC for presentation.
			\emph{Below, left to right:} Plots for each summarization dataset
           showing increasing diversity of summaries along both dimensions of extraction in \DATASET.
           }
		\label{figure:histograms}
	\end{figure*}

}

\newcommand{\TABLEDATASETS}{

	\begin{table*}
		\centering
        \small

		\begin{tabular}{l rrr | rrr | rrr rrr}
			&\multicolumn{3}{c}{\textsc{DUC 2003 \& 2004}}
			&\multicolumn{3}{c}{\textsc{CNN / Daily Mail}}
			&\multicolumn{3}{c}{\textsc{Newsroom - T}}
			&\multicolumn{3}{c}{\textsc{Newsroom - U}}
			
			\\
			\cmidrule(lr){2-4}
			\cmidrule(lr){5-7}
			\cmidrule(lr){8-10}
			\cmidrule(lr){11-13}
			& \multicolumn{1}{c}{R-1}
			& \multicolumn{1}{c}{R-2}
			& \multicolumn{1}{c}{R-L}
			& \multicolumn{1}{c}{R-1}
			& \multicolumn{1}{c}{R-2}
			& \multicolumn{1}{c}{R-L}
			& \multicolumn{1}{c}{R-1}
			& \multicolumn{1}{c}{R-2}
			& \multicolumn{1}{c}{R-L}
			& \multicolumn{1}{c}{R-1}
			& \multicolumn{1}{c}{R-2}
			& \multicolumn{1}{c}{R-L}
			\\ \midrule
			Lede-3    &    12.99  &    3.89  &    11.44  &    38.64  &    17.12  &    35.13  &    30.49  &    21.27  &    28.42  &    30.63  &    21.41  &    28.57  \\
			Fragments &    87.04  &   68.45  &    87.04  &    93.36  &    83.19  &    93.36  &    88.46  &    76.03  &    88.46  &    88.48  &    76.06  &    88.48  \\ \midrule
			TextRank  &    15.75  &    4.06  &    13.02  &    29.06  &    11.14  &    24.57  &    22.77  &     9.79  &    18.98  &    22.76  &     9.80  &    18.97  \\
			Abs-N     &     2.44  &    0.04  &     2.37  &     5.07  &     0.16  &     4.80  &     5.88  &     0.39  &     5.32  &     5.90  &     0.43  &     5.36  \\
			Pointer-C &    12.40  &    2.88  &    10.74  &    32.51  &    11.90  & \x{28.95} &    20.25  &     7.32  &    17.30  &    20.29  &     7.33  &    17.31  \\ 
			Pointer-S &    15.10  &    4.55  &    12.42  & \x{34.33} & \x{13.79} &    28.42  &    24.50  &    12.60  &    20.33  &    24.48  &    12.52  &    20.30  \\ 
			Pointer-N & \y{17.29} & \y{5.01} & \y{14.53} &    31.61  &    11.70  &    27.23  & \x{26.02} & \x{13.25} & \x{22.43} & \x{26.04} & \x{13.24} & \x{22.45} \\ \midrule
		\end{tabular}

		\caption{
		ROUGE-1, ROUGE-2, and ROUGE-L scores for baselines and systems on two common existing datasets, the combined DUC~2003 \& 2004 datasets and CNN~/~Daily~Mail dataset, and the released (T) and unreleased (U) test sets of \DATASET.
        The best results for non-baseline systems in the lower parts of the table are in \textbf{bold}.
	    \vspace{0.5em}
		}
		\label{table:datasets}
		
	\end{table*}

}

\newcommand{\TABLESUBSETS}{

	\begin{table*}
		\centering
        \small

		\begin{tabular}{l rrr rrr rrr | rrr}
			&\multicolumn{3}{c}{\textsc{Extractive}}
			&\multicolumn{3}{c}{\textsc{Mixed}}
			&\multicolumn{3}{c}{\textsc{Abstractive}}
			&\multicolumn{3}{c}{\textsc{Newsroom - D}}
			\\
			\cmidrule(lr){2-4}
			\cmidrule(lr){5-7}
			\cmidrule(lr){8-10}
			\cmidrule(lr){11-13}
			& \multicolumn{1}{c}{R-1}
			& \multicolumn{1}{c}{R-2}
			& \multicolumn{1}{c}{R-L}
			& \multicolumn{1}{c}{R-1}
			& \multicolumn{1}{c}{R-2}
			& \multicolumn{1}{c}{R-L}
			& \multicolumn{1}{c}{R-1}
			& \multicolumn{1}{c}{R-2}
			& \multicolumn{1}{c}{R-L}
			& \multicolumn{1}{c}{R-1}
			& \multicolumn{1}{c}{R-2}
			& \multicolumn{1}{c}{R-L}
			\\ \midrule
			Lede-3      &    53.05  &    49.01  &    52.37  &    25.15  &    12.88  &    22.08  &    13.69  &     2.42  &    11.24  &    30.72  &    21.53  &    28.65 \\
			Fragments   &    98.95  &    97.89  &    98.95  &    92.68  &    82.09  &    92.68  &    73.43  &    47.66  &    73.43  &    88.46  &    76.07  &    88.46 \\ \midrule
			TextRank    &    32.43  &    19.68  &    28.68  &    22.30  &     7.87  &    17.75  &    13.54  &     1.88  &    10.46  &    22.82  &     9.85  &    19.02 \\
			Abs-N       &     6.08  &     0.21  &     5.42  &     5.67  &     0.15  &     5.08  &     6.21  &     1.07  &     5.68  &     5.98  &     0.48  &     5.39 \\
			Pointer-C   &    28.34  &    14.65  &    25.21  &    20.22  &     6.51  &    16.88  &    13.11  &     1.62  &    10.72  &    20.47  &     7.50  &    17.51 \\
			Pointer-S   &    37.29  &    26.56  &    33.34  &    23.71  &    10.59  &    18.79  &    13.89  &     2.22  &    10.34  &    24.83  &    12.94  &    20.66 \\
			Pointer-N   & \x{39.11} & \x{27.95} & \x{36.17} & \y{25.48} & \x{11.04} & \x{21.06} & \y{14.66} &  \x{2.26} & \y{11.44} & \x{26.27} & \x{13.55} & \x{22.72} \\ \midrule
		\end{tabular}

		\caption{
		Performance of the baselines and systems on the three extractiveness subsets of the \DATASET development set, and the overall scores of systems on the full development set (D). The best results for non-baseline systems in the lower parts of the table are in \textbf{bold}.
		}
		\label{table:subsets}
	\end{table*}

}

\newcommand{\TABLESUBSETSCOMPRESSION}{

	\begin{table*}[bp]
		\centering
        \small

		\begin{tabular}{l rrr rrr rrr}
			&\multicolumn{3}{c}{\textsc{Low Compression}}
			&\multicolumn{3}{c}{\textsc{Medium}}
			&\multicolumn{3}{c}{\textsc{High Compression}}
			\\
			\cmidrule(lr){2-4}
			\cmidrule(lr){5-7}
			\cmidrule(lr){8-10}
			& \multicolumn{1}{c}{R-1}
			& \multicolumn{1}{c}{R-2}
			& \multicolumn{1}{c}{R-L}
			& \multicolumn{1}{c}{R-1}
			& \multicolumn{1}{c}{R-2}
			& \multicolumn{1}{c}{R-L}
			& \multicolumn{1}{c}{R-1}
			& \multicolumn{1}{c}{R-2}
			& \multicolumn{1}{c}{R-L}
			\\ \midrule
			Lede-3      &    42.89  &    34.91  &    41.06  &    30.62  &    20.77  &    28.30  &    18.57  &     8.83  &    16.53  \\
			Fragments   &    87.78  &    77.20  &    87.78  &    89.73  &    77.66  &    89.73  &    87.88  &    73.34  &    87.88  \\ \midrule
			TextRank    &    30.35  &    17.51  &    26.67  &    22.98  &     8.69  &    18.56  &    15.07  &     3.31  &    11.78  \\
			Abs-N       &     6.27  &     0.75  &     5.65  &     6.22  &     0.52  &     5.60  &     5.48  &     0.18  &     4.93  \\
			Pointer-C   &    27.47  &    13.49  &    24.18  &    20.05  &     6.25  &    16.76  &    14.07  &     2.89  &    11.76  \\
			Pointer-S   &    35.42  &    23.43  &    30.89  &    24.11  &    11.28  &    19.45  &    15.31  &     4.46  &    11.98  \\
			Pointer-N   & \x{36.96} & \x{24.52} & \x{33.43} & \x{25.56} & \x{11.68} & \x{21.47} & \x{16.57} &  \x{4.72} & \x{13.52} \\ \midrule
		\end{tabular}

		\caption{
            Performance of the baselines and systems on the three compression subsets of the \DATASET development set.
            Article-summary pairs with low compression have longer reference summaries with respect to their texts.
            Article-summary pairs with high compression have shorter reference summaries with respect to their texts.
            \vspace{5em}
		}
		\label{table:compressionsubsets}
	\end{table*}

}

\newcommand{\TABLESUBSETSCOVERAGE}{

	\begin{table*}[bp]
		\centering
        \small

		\begin{tabular}{l rrr rrr rrr}
			&\multicolumn{3}{c}{\textsc{Low Coverage}}
			&\multicolumn{3}{c}{\textsc{Medium}}
			&\multicolumn{3}{c}{\textsc{High Coverage}}
			\\
			\cmidrule(lr){2-4}
			\cmidrule(lr){5-7}
			\cmidrule(lr){8-10}
			& \multicolumn{1}{c}{R-1}
			& \multicolumn{1}{c}{R-2}
			& \multicolumn{1}{c}{R-L}
			& \multicolumn{1}{c}{R-1}
			& \multicolumn{1}{c}{R-2}
			& \multicolumn{1}{c}{R-L}
			& \multicolumn{1}{c}{R-1}
			& \multicolumn{1}{c}{R-2}
			& \multicolumn{1}{c}{R-L}
			\\ \midrule
			
			Lede-3      &    15.07  &     4.02  &    12.66  &    29.66  &    18.69  &    26.98  &    46.89  &    41.25  &    45.77  \\
			Fragments   &    72.45  &    46.16  &    72.45  &    93.41  &    83.08  &    93.41  &    99.13  &    98.16  &    99.13  \\ \midrule
			TextRank    &    14.43  &     2.80  &    11.36  &    23.62  &     9.48  &    19.27  &    30.15  &    17.04  &    26.18  \\
			Abs-N       &     6.25  &     1.09  &     5.72  &     5.61  &     0.15  &     5.05  &     6.10  &     0.19  &     5.40  \\
			Pointer-C   &    13.99  &     2.46  &    11.57  &    21.70  &     8.06  &    18.47  &    25.80  &    12.06  &    22.57  \\
			Pointer-S   &    15.16  &     3.63  &    11.61  &    26.95  &    14.51  &    22.30  &    32.42  &    20.77  &    28.15  \\
			Pointer-N   & \y{16.07} &  \x{3.78} & \y{12.85} & \x{28.79} & \x{15.31} & \x{24.79} & \x{34.03} & \x{21.67} & \x{30.62} \\ \midrule
		\end{tabular}

		\caption{
            Performance of the baselines and systems on the three coverage subsets of the \DATASET development set.
            Article-summary pairs with low coverage have reference summaries that borrow words less frequently from their texts and contain more novel words and phrases.
            Article-summary pairs with high coverage borrow more words from their text and include fewer novel words and phrases.
            \vspace{5em}
		}
		\label{table:coveragesubsets}
	\end{table*}

}

\newcommand{\TABLEHUMAN}{

	\begin{table}[t]

		\centering
        \small

		\begin{tabular}{ l r r r r | r}
			& \multicolumn{2}{c}{~~\textsc{Semantic}}
			& \multicolumn{2}{c}{~~\textsc{Syntactic}}
			\\
			\cmidrule(l){2-3}
			\cmidrule(l){4-5}
			& \multicolumn{1}{c}{\textsc{INF}}
            & \multicolumn{1}{c}{\textsc{REL}}
			& \multicolumn{1}{c}{\textsc{FLU}}
			& \multicolumn{1}{c}{\textsc{COH}}
			& \multicolumn{1}{c}{Avg.}
			\\ \midrule
			Lede-3      &    3.98  &    4.13  &    4.13  &    4.08  &    4.08  \\
			Fragments   &    2.91  &    3.26  &    3.09  &    3.06  &    3.08  \\ \midrule
			TextRank    &    3.61  &    3.92  & \x{3.87} & \x{3.86} & \x{3.81} \\
			Abs-N       &    2.09  &    2.35  &    2.66  &    2.50  &    2.40  \\
			Pointer-C   &    3.55  &    3.78  &    3.22  &    3.30  &    3.46  \\
			Pointer-S   & \x{3.77} & \x{4.02} &    3.56  &    3.56  &    3.73  \\
			Pointer-N   &    3.36  &    3.82  &    3.43  &    3.39  &    3.50  \\ \midrule
		\end{tabular}

		\caption{
            Average performance of systems as scored by human evaluators.
            Each summary was scored by three different evaluators.
            \emph{Dimensions, from left to right:} informativeness, relevance, fluency, and coherence,
            and a mean of the four dimensions for each system.
		}
		\label{table:human}

	\end{table}

}

\newcommand{\TABLEPROMPT}{

	\begin{table}[t]

		\centering
        \small

		\begin{tabular}{ l p{15.5em} }
			\textsc{Dimension} & \textsc{Prompt} \\ \midrule
			Informativeness    & How well does the summary capture the key points of the article?                  \vspace{0.3em} \\
			Relevance          & Are the details provided by the summary consistent with details in the article?   \vspace{0.3em} \\
			Fluency            & Are the individual sentences of the summary well-written and grammatical?         \vspace{0.3em} \\
			Coherence          & Do phrases and sentences of the summary fit together and make sense collectively? \\ \midrule
		\end{tabular}

		\caption{
			The prompts given to Amazon Mechanical Turk crowdworkers for evaluating each summary.
			\vspace{0.5em}
		}
		\label{table:crowdquestions}

	\end{table}

}

%% file: sections/0abstract.tex
\begin{abstract}

We present \DATASET, a summarization dataset of 1.3~million articles and summaries written by authors and editors in newsrooms of 38 major news publications.
Extracted from search and social media metadata between 1998 and 2017, these high-quality summaries demonstrate high diversity of summarization styles.
In particular, the summaries combine \emph{abstractive} and \emph{extractive} strategies, borrowing words and phrases from articles at varying rates.
We analyze the extraction strategies used in \DATASET summaries against other datasets to quantify the diversity and difficulty of our new data, and train existing methods on the data to evaluate its utility and challenges. 

\end{abstract}

%% file: sections/1introduction.tex
\section{Introduction}
\label{section:introduction}

The development of learning methods for automatic summarization is constrained by the limited high-quality data available for training and evaluation.
Large datasets have driven rapid improvement in other natural language generation tasks, such as machine translation, where data size and diversity have proven critical for modeling the alignment between source and target texts~\cite{Tiedemann2012}.
Similar challenges exist in summarization, with the additional complications introduced by the length of source texts and the diversity of summarization strategies used by writers.
Access to large-scale high-quality data is an essential prerequisite for making substantial progress in summarization.
In this paper, we present \mbox{\DATASET}, a dataset with 1.3 million news articles and human-written summaries. 

\SUBSETFIGURE

\DATASET's summaries were written by authors and editors in the newsrooms of news, sports, entertainment, financial, and other publications.
The summaries were published with articles as HTML metadata for social media services and search engines page descriptions.
\mbox{\DATASET} summaries are written by humans, for common readers, and with the explicit purpose of summarization.
As a result, \DATASET is a nearly two decade-long snapshot representing how single-document summarization is used in practice across a variety of sources, writers, and topics.

Identifying large, high-quality resources for summarization has called for creative solutions in the past. 
This includes using news headlines as summaries of article prefixes~\cite{Napoles2012, Rush2015}, concatenating bullet points as summaries~\cite{Hermann2015, See2017}, or using librarian archival summaries~\cite{NYTimesCorpus2008}. 
While these solutions provide large scale data, it comes at the cost of how well they reflect the summarization problem or their focus on very specific styles of summarizations, as we discuss in Section~\ref{section:analysis}. 
\DATASET is distinguished from these resources in its combination of size and diversity. 
The summaries were written with the explicit goal of concisely summarizing news articles over almost two decades.
Rather than rely on a single source, the dataset includes summaries from 38 major publishers. 
This diversity of sources and time span translate into a diversity of summarization styles.

We explore \DATASET to better understand the dataset and how summarization is used in practice by newsrooms.
Our analysis focuses on a key dimension, \emph{extractivenss} and \emph{abstractiveness}: extractive summaries frequently borrow words and phrases from their source text, while abstractive summaries describe the contents of articles primarily using new language.
We develop measures designed to quantify extractiveness and use these measures to subdivide the data into extractive, mixed, and abstractive subsets, as shown in Figure~\ref{figure:examples}, displaying the broad set of summarization techniques practiced by different publishers. 

Finally, we analyze the performance of three summarization models as baselines for \nb{\DATASET} to better understand the challenges the dataset poses.
In addition to automated ROUGE evaluation~\cite{Lin2004b, Lin2004}, we design and execute a  benchmark human evaluation protocol to quantify the output summaries relevance and quality.
Our experiments demonstrate that \DATASET presents an open challenge for summarization systems, while providing a large resource to enable data-intensive learning methods.  
The dataset and evaluation protocol are available online at \WEBSITE.

%% file: sections/2related.tex
\section{Existing Datasets}
\label{section:datasets}

There are a several frequently used summarization datasets.
Listed in Figure~\ref{figure:datasets} are examples from four datasets.
The examples are chosen to be representative: they have scores within 5\% of their dataset average across our analysis measures (Section~\ref{section:analysis}). 
To illustrate the extractive and abstractive nature of summaries, we underline \ulb{multi-word phrases} shared between the article and summary, and italicize \emph{words} used only in the summary.

\FIGUREEXAMPLESUMMARIES

\subsection{Document Understanding Conference}
\label{section:datasets:duc}

Datasets produced for the Document Understanding Conference (DUC)\footurl{http://duc.nist.gov/} are small, high-quality datasets developed to evaluate summarization systems \cite{Harman2004, Dang2006}.

DUC data consist of newswire articles paired with human summaries written specifically for DUC.
One distinctive feature of the DUC datasets is the availability of multiple reference summaries for each article.
This is a major advantage of DUC compared to other datasets, especially when evaluating with ROUGE~\cite{Lin2004, Lin2004b}, which was designed to be used with multiple references.
However, DUC datasets are small, which makes it difficult to use them as training data. 

DUC summaries are often used in conjunction with larger training datasets, including Gigaword~\cite{Rush2015, Chopra2016}, CNN~/~Daily~Mail~\cite{Nallapati2017, Paulus2017, See2017}, or Daily Mail alone~\cite{Nallapati2016, Cheng2016}. 
The data have also been used to evaluate unsupervised methods~\cite{Dorr2003, Mihalcea04, Barrios2016}.

\subsection{Gigaword}
\label{section:datasets:gigaword}

The Gigaword Corpus~\cite{Napoles2012} contains nearly 10 million documents from seven  newswire sources, including the Associated Press, New York Times Newswire Service, and Washington Post Newswire Service.
Compared to other existing datasets used for summarization, the Gigaword corpus is the largest and most diverse in its sources.
While Gigaword does not contain summaries, prior work uses Gigaword headlines as simulated summaries~\cite{Rush2015, Chopra2016}.
These systems are trained on Gigaword to recreate headlines given the first sentence of an article.
When used this way, Gigaword's simulated summaries are shorter than most natural summary text.
Gigaword, along with similar text-headline datasets~\cite{Filippova2013}, are also used for the related sentence compression task~\cite{Dorr2003,Filippova2015}. 

\subsection{New York Times Corpus}
\label{section:datasets:nytimes}

The New York Times Annotated Corpus~\cite{NYTimesCorpus2008} is the largest summarization dataset currently available.
It consists of carefully curated articles from a single source, The New York Times.
The corpus contains several hundred thousand articles written between 1987--2007 that have paired summaries.
The summaries were written for the corpus by library scientists, rather than at the time of publication.
Our analysis in Section~\ref{section:analysis} reveals that the data are somewhat biased toward extractive strategies, making it particularly useful as an extractive summarization dataset.
Despite this, limited work has used this dataset for summarization~\cite{Hong2014, Durrett2016, Paulus2017}.

\subsection{CNN / Daily Mail}
\label{section:datasets:cnndm}

The CNN~/~Daily~Mail question answering dataset~\cite{Hermann2015} is frequently used for summarization.
The dataset includes CNN and Daily Mail articles, each associated with several bullet point descriptions. 
When used in summarization, the bullet points are typically concatenated into a single summary.\footurl{https://github.com/abisee/cnn-dailymail}
The dataset has been used for summarization as is~\cite{See2017}, or after pre-processing for entity anonymization~\cite{Nallapati2017}. 
This different usage makes comparisons between systems using these data challenging.
Additionally, some systems use both CNN and Daily Mail for training~\cite{Nallapati2017, Paulus2017, See2017}, whereas others use only Daily Mail articles~\cite{Nallapati2016, Cheng2016}.
Our analysis shows that the CNN~/~Daily~Mail summaries have strong bias toward extraction (Section~\ref{section:analysis}).
Similar observations about the data were made by \citet{Chen2016} with respect to the question answering task.

%% file: sections/3collection.tex
\section{Collecting \DATASET Summaries}
\label{section:collection}

The \DATASET dataset was collected using social media and search engine metadata.
To create the dataset, we performed a Web-scale crawling of over 100 million pages from a set of online publishers.
We identify newswire articles and use the summaries provided in the HTML metadata. These summaries were created to be used in search engines and social media.

We collected HTML pages and metadata using the Internet Archive (\href{https://archive.org/}{Archive.org}), accessing archived pages of a large number of popular news, sports, and entertainment sites.
Using Archive.org provides two key benefits.
First, the archive provides an API that allows for collection of data across time, not limited to recently available articles.
Second, the archived URLs of the dataset articles are immutable, allowing distribution of this dataset using a thin, URL-only list.

The publisher sites we crawled were selected using a combination of Alexa.com top overall sites, as well as Alexa's top news sites.\footnote{Alexa removed the extended public list in 2017, see: \\ \url{https://web.archive.org/web/2016/https://www.alexa.com/topsites/category/News}}
We supplemented the lists with older lists published by Google of the highest-traffic sites on the Web.\footnote{Google removed this list in 2013, see: \\ \url{https://web.archive.org/web/2012/http://www.google.com/adplanner/static/top1000}}
We excluded sites such as Reddit that primarily aggregate rather than produce content, as well as publisher sites that proved to have few or no articles with summary metadata available, or have articles primarily in languages other than English.
This process resulted in a set of 38 publishers that were included in the dataset.

\subsection{Content Scraping}
\label{section:collection:scraping}

We used two techniques to identify article pages from the selected publishers on Archive.org: the search API and index-page crawl. The API allows queries using URL pattern matching, which focuses article crawling on high-precision subdomains or paths.
We used the API to search for content from the publisher domains, using specific patterns or post-processing filtering to ensure article content. 
In addition, we used Archive.org to retrieve the historical versions of the home page for all publisher domains.
The archive has content from 1998 to 2017 with varying degrees of time resolution.
We obtained at least one snapshot of each page for every available day.
For each snapshot, we retrieved all articles listed on the page.

For both search and crawled URLs, we performed article de-duplication using URLs to control for varying URL fragments, query parameters, protocols, and ports.
When performing the merge, we retained only the earliest article version available to prevent the collection of stale summaries that are not updated when articles are changed.

\subsection{Content Extraction}
\label{section:collection:extraction}

Following identification and de-duplication, we extracted the article texts and summaries and further cleaned and filtered the dataset. 

\paragraph{Article Text}

We used Readability\footurl{https://pypi.org/project/readability-lxml/0.6.2/} to extract HTML body content.
Readability uses HTML heuristics to extract the main content and title of a page, producing article text without extraneous HTML markup and images.
Our preliminary testing, as well as comparison by \citet{Peters2015}, found Readability to be one of the highest accuracy content extraction algorithms available.
To exclude inline advertising and image captions sometimes present in extractions, we applied additional filtering of paragraphs with fewer than five words.
We excluded articles with no body text extracted.

\paragraph{Summary Metadata}

We extracted the article summaries from the metadata available in the HTML pages of articles.
These summaries are often written by newsroom editors and journalists to appear in social media distribution and search results. 
While there is no standard metadata format for summaries online, common fields are often present in the page's HTML.
Popular metadata field types include:
\textit{og:description},
\textit{twitter:description},
and \textit{description}. 
In cases where different metadata summaries were available, and were different, we used the first field available according to the order above. 
We excluded articles with no summary text of any type. 
We also removed article-summary pairs with a high amount of precisely-overlapping text to remove rule-based automatically-generated summaries fully copied from the article (e.g., the first paragraph). 

\subsection{Building the Dataset}
\label{section:collection:overview}

Our scraping and extraction process resulted in a set of 1,321,995 article-summary pairs. 
Simple dataset statistics are shown in Table~\ref{table:statistics}.
The data are divided into training~(76\%), development~(8\%), test~(8\%), and unreleased test~(8\%) datasets using a hash function of the article URL.
We use the articles' Archive.org URLs for lightweight distribution of the data.
Archive.org is an ideal platform for distributing the data, encouraging its users to scrape its resources.
We provide the extraction and analysis scripts used during data collection for reproducing the full dataset from the URL list.

\OVERVIEWTABLE

%% file: sections/4analysis.tex
\section{Data Analysis}
\label{section:analysis}

\DATASET contains summaries from different topic domains, written by many authors, over the span of more than two decades. 
This diversity is an important aspect of the dataset.
We analyze the data to quantify the differences in summarization styles and techniques between the different publications to show the importance of reflecting this diversity. 
In Sections~\ref{section:evaluation} and~\ref{section:evaluation:human}, we examine the effect of the dataset diversity on the performance of a variety of summarization systems.

\subsection{Characterizing Summarization Strategies}
\label{section:analysis:dimensions}

\FRAGMENTSALGORITHM

We examine summarization strategies using three measures that capture the degree of text overlap between the summary and article, and the rate of compression of the information conveyed.

Given an article text $A~=~\langle a_1, a_2, \ldots, a_n \rangle$ consisting of a sequence of tokens $a_i$ and the corresponding article summary $S~=~\langle s_1, s_2, \cdots, s_m \rangle$ consisting of tokens $s_i$, the set of extractive fragments $\mathcal{F}(A, S)$ is the set of shared sequences of tokens in~$A$~and~$S$.
We identify these extractive fragments of an article-summary pair using a greedy process. 
We process the tokens in the summary in order. At each position, if there is a sequence of tokens in the source text that is prefix of the remainder of the summary, we mark this prefix as extractive and continue. We prefer to mark the longest prefix possible at each step. Otherwise, we mark the current summary token as abstractive. The set $\mathcal{F}(A, S)$ includes all the tokens sequences identified as extractive. 
Figure~\ref{figure:fragments} formally describes this procedure. 
Underlined phrases of Figures~\ref{figure:examples} and~\ref{figure:datasets} are examples of fragments identified as extractive.
Using $\mathcal{F}(A, S)$, we compute two measures: \emph{extractive fragment coverage} and \emph{extractive fragment density}.

\paragraph{Extractive Fragment Coverage}

The coverage measure quantifies the extent to which a summary is derivative of a text.
$\textsc{Coverage}(A, S)$ measures the percentage of words in the summary that are part of an extractive fragment with the article:

\vspace{0.3em}
\begin{small}
\begin{equation*}
	\textsc{Coverage}(A, S) = \frac{1}{|S|} \sum\limits_{f \in \mathcal{F}(A, S)} |f|\;\;.
\end{equation*}
\end{small}

\noindent
For example, a summary with 10 words that borrows 7 words from its article text and includes 3 new words will have $\textsc{Coverage}(A, S) = 0.7$.

\paragraph{Extractive Fragment Density}

The density measure quantifies how well the word sequence of a summary can be described as a series of extractions.
For instance, a summary might contain many individual words from the article and therefore have a high coverage.
However, if arranged in a new order, the words of the summary could still be used to convey ideas not present in the article.
We define $\textsc{Density}(A, S)$ as the average length of the extractive fragment to which each word in the summary belongs.
The density formulation is similar to the coverage definition but uses a square of the fragment length:

\vspace{0.3em}
\begin{small}
\begin{equation*}
	\textsc{Density}(A, S) = \frac{1}{|S|} \sum\limits_{f \in \mathcal{F}(A, S)} |f|^2\;\;.
\end{equation*}
\end{small}

\noindent
For example, an article with a 10-word summary made of two extractive fragments of lengths 3 and 4 would have $\textsc{Coverage}(A, S) = 0.7$ and $\textsc{Density}(A, S) = 2.5$.

\paragraph{Compression Ratio}

We use a simple dimension of summarization, \emph{compression ratio}, to further characterize summarization strategies.
We define \textsc{Compression} as the word ratio between the article and summary:

\vspace{0.5em}
\begin{small}
\begin{equation*}
	\textsc{Compression}(A, S) = |A| ~\big/~ |S|\;\;.
\end{equation*}
\end{small}

\vspace{-0.5em}

\noindent
Summarizing with higher compression is challenging as it requires capturing more precisely the critical aspects of the article text.

\subsection{Analysis of Dataset Diversity}
\label{section:analysis:diversity}

\HISTOGRAMFIGURE

We use density, coverage, and compression to understand the distribution of human summarization techniques across different sources. 
Figure~\ref{figure:histograms} shows the distributions of summaries for different domains in the \DATASET dataset, along with three major existing summarization datasets: DUC 2003-2004 (combined), CNN~/~Daily Mail, and the New York Times Corpus.

\paragraph{Publication Diversity}

Each \DATASET publication shows a unique distribution of summaries mixing extractive and abstractive strategies in varying amounts.
For example, the third entry on the top row shows the summarization strategy used by BuzzFeed.
The density (y-axis) is relatively low, meaning BuzzFeed summaries are unlikely to include long extractive fragments.
While the coverage (x-axis) is more varied, BuzzFeed's coverage tends to be lower, indicating that it frequently uses novel words in summaries.
The publication plots in the figure are sorted by median compression ratio.
We observe that publications with lower compression ratio (top-left of the figure) exhibit higher diversity along both dimensions of extractiveness.
However, as the median compression ratio increases, the distributions become more concentrated, indicating that summarization strategies become more rigid.

\paragraph{Dataset Diversity}

Figure~\ref{figure:histograms} demonstrates how DUC, CNN~/~Daily~Mail, and the New York Times exhibit different human summarization strategies.
DUC summarization is fairly similar to the high-compression newsrooms shown in the lower publication plots in  Figure~\ref{figure:histograms}.
However, DUC's median compression ratio is much higher than all other datasets and \DATASET publications.
The figure shows that CNN~/~Daily Mail and New York Times are skewed toward extractive summaries with lower compression ratios. 
CNN~/~Daily Mail shows higher coverage and density than all other datasets and publishers in our data.
Compared to existing datasets, \DATASET covers a much larger range of summarization styles, ranging from both highly extractive to highly abstractive.

%% file: sections/5methods.tex
\section{Performance of Existing Systems}
\label{section:systems}

We train and evaluate several summarization systems to understand the challenges of \DATASET and its usefulness for training systems. 
We evaluate three systems, each using a different summarization strategy with respect to extractiveness:
fully extractive (\TextRank),
fully abstractive (Seq2Seq),
and mixed (pointer-generator).
We further study the performance of the pointer-generator model on \DATASET by training three systems using different dataset configurations. 
We compare these systems to two rule-based systems that provide baseline (Lede-3) and an extractive oracle (Fragments). 

\paragraph{Extractive: TextRank}

\TextRank is a sentence-level extractive summarization system. 
The system was originally developed by \citet{Mihalcea04} and was later further developed and improved by \citet{Barrios2016}. 
\TextRank uses an unsupervised sentence-ranking approach similar to Google PageRank~\cite{Page1999}.
\TextRank picks a sequence of sentences from a text for the summary up to a maximum allowable length.
While this maximum length is typically preset by the user, in order to optimize ROUGE scoring, we tune this parameter to optimize ROUGE-1 $F_1$-score on the \DATASET training data.
We experimented with values between 1--200, and found the optimal value to be 50 words.
We use tuned \TextRank of in Tables~\ref{table:datasets},~\ref{table:subsets}, and in the supplementary material.

\paragraph{Abstractive: Seq2Seq / Attention}

Sequence-to-sequence models with attention~\cite{Cho2014,Sutskever2014,Bahdanau2014} have been applied to various language tasks, including summarization~\cite{Chopra2016, Nallapati2016b}. 
The process by which the model produces tokens is abstractive, as there is no explicit mechanism to copy tokens from the input text.
We train a TensorFlow implementation\footurl{https://github.com/tensorflow/models/tree/f87a58/research/textsum} of the \citet{Rush2015} model using \DATASET.

\paragraph{Mixed: Pointer-Generator}

The pointer-generator model~\cite{See2017} uses abstractive token generation and extractive token copying using a pointer mechanism~\cite{Vinyals2015, Glehre2016}, keeping track of extractions using coverage \cite{Tu2016}.
We evaluate three instances of this model by varying the training data: (1) Pointer-C: trained on the CNN~/~Daily~Mail dataset; (2) Pointer-N: trained on the \DATASET dataset; and (3) Pointer-S: trained on a random subset of \DATASET training data the same size as the CNN~/~Daily~Mail training. 
The last instance aims to understand the effects of dataset size and summary diversity.

\paragraph{Lower Bound: Lede-3}

A common automatic summarization strategy of online publications is to copy the first sentence, first paragraph, or first~$k$~words of the text and treat this as the summary.
Following prior work~\cite{See2017, Nallapati2017}, we use the Lede-3 baseline, in which the first three sentences of the text are returned as the summary.
Though simple, this baseline is competitive with state-of-the-art systems.

\paragraph{Extractive Oracle: Fragments}

This system has access to the reference summary.
Given an article $A$ and its summary $S$, the system computes $\mathcal{F}(A, S)$ (Section~\ref{section:analysis}).
Fragments concatenates the fragments in $\mathcal{F}(A, S)$ in the order they appear in the summary, representing the best possible performance of an ideal extractive system.
Only systems that are capable of abstractive reasoning can outperform the ROUGE scores of Fragments.

%% file: sections/6evaluation.tex
\section{Automatic Evaluation}
\label{section:evaluation}

\TABLEDATASETS
\TABLESUBSETS

We study model performance of \DATASET, CNN~/~Daily~Mail, and the combined DUC 2003 and 2004 datasets. We use the five systems described in Section~\ref{section:systems}, including the extractive oracle. 
We also evaluate the systems using subsets of \DATASET to characterize the sensitivity of systems to different levels of extractiveness in reference summaries.
We use the $F_1$-score variants of ROUGE-1, ROUGE-2, and ROUGE-L to account for different summary lengths.
ROUGE scores are computed with the default configuration of the \citet{Lin2004} ROUGE v1.5.5 reference implementation.
Input article text and reference summaries for all systems are tokenized using the Stanford CoreNLP tokenizer~\cite{Manning2014}.

Table~\ref{table:datasets} shows results for summarization systems on DUC, CNN~/~Daily~Mail, and \DATASET.
In nearly all cases, the fully extractive Lede-3 baseline produces the most successful summaries, with the exception of the relatively extractive DUC.
Among models, \DATASET-trained Pointer-N performs best on all datasets other than CNN~/~Daily~Mail, an out-of-domain dataset.
Pointer-C, which has access to only a limited subset of \DATASET, performs worse than Pointer-N on average.
However, despite not being trained on CNN~/~Daily~Mail, Pointer-S outperforms Pointer-C on its own data under ROUGE-N and is competitive under ROUGE-L.
Finally, both Pointer-N and Pointer-S outperform other systems and baselines on DUC, whereas Pointer-C does not outperform Lede-3.

Table~\ref{table:subsets} shows development results on the \DATASET data for different level of extractiveness. 
Pointer-N outperforms the remaining models across all extractive subsets of \DATASET and, in the case of the abstractive subset, exceeds the performance of Lede-3.
The success of Pointer-N and Pointer-S in generalizing and outperforming models on DUC and CNN~/~Daily~Mail indicates the usefulness of \DATASET in generalizing to out-of-domain data.
Similar subset analysis for our other two measures, coverage and compression, are included in the supplementary material.

\section{Human Evaluation}
\label{section:evaluation:human}

ROUGE scores systems using frequencies of shared $n$-grams.
Evaluating systems with ROUGE alone biases scoring against abstractive systems, which rely more on paraphrasing.
To overcome this limitation, we provide human evaluation of the different systems on \DATASET.
While human evaluation is still uncommon in summarization work, developing a benchmark dataset presents an opportunity for developing an accompanying protocol for human evaluation.

Our evaluation method is centered around three objectives:
(1) distinguishing between syntactic and semantic summarization quality,
(2) providing a reliable (consistent and replicable) measurement, and
(3) allowing for portability such that the measure can be applied to other models or summarization datasets.

We select two semantic and two syntactic dimensions for evaluation based on experiments with evaluation tasks by \citet{Paulus2017} and \citet{Tan2017}. 
The two semantic dimensions, summary \emph{informativeness} (INF) and \emph{relevance} (REL), measure whether the system-generated text is useful as a summary, and appropriate for the source text, respectively.
The two syntactic dimensions, \emph{fluency} (FLU) and \emph{coherence} (COH), measure whether individual sentences or phrases of the summary are well-written and whether the summary as a whole makes sense respectively.
Evaluation was performed on 60 summaries, 20 from each extractive \DATASET subset.
Each system-article pair was evaluated by three unique raters.
Exact prompts given to raters for each dimension are shown in Table~\ref{table:crowdquestions}.

\TABLEPROMPT
\TABLEHUMAN

Table~\ref{table:human} shows the mean score given to each system under each of the four dimensions, as well as the mean overall score (rightmost column).
No summarization system exceeded the scores given to the Lede-3 baseline.
However, the extractive oracle designed to maximize $n$-gram based evaluation performed worse than the majority of systems under human evaluation.
While the fully abstractive Abs-N model performed very poorly under automatic evaluation, it fared slightly better when scored by humans.
\TextRank received the highest overall score.
\TextRank generates full sentences extracted from the article, and raters preferred \TextRank primarily for its fluency and coherence.
The pointer-generator models do not have this advantage, and raters did not find the pointer-generator models to be as syntactically sound as \TextRank.
However, raters preferred the informativeness and relevance of the \nb{Pointer-S} and Pointer-N models, though not the Pointer-C model, over \TextRank.

%% file: sections/7conclusion.tex
\section{Conclusion}
\label{section:conclusion}

We present \DATASET, a dataset of articles and their summaries written in the newsrooms of online publications.
\DATASET is the largest summarization dataset available to date, and exhibits a wide variety of human summarization strategies.
Our proposed measures and the analysis of strategies used by different publications and articles propose new directions for evaluating the difficulty of summarization tasks and for developing future summarization models.
We show that the dataset's diversity of summaries presents a new challenge to summarization systems. 
Finally, we find that using \DATASET to train an existing state-of-art mixed-strategy summarization model results in performance improvements on out-of-domain data.
The \DATASET dataset is available online at \WEBSITE.

\section*{Acknowledgements}
\label{section:acknowledgements}

This work is funded by Oath as part of the Connected Experiences Laboratory and by a Google Research Award. 
We thank the anonymous reviewers for their feedback.

%% file: sections/appendix.tex
\appendix

\vfill\null
\pagebreak

\section*{Additional Evaluation}

In Section~\ref{section:analysis}, we discuss three measures of summarization diversity: coverage, density, and compression.
In addition to quantifying diversity of summarization strategies, these measures are helpful for system error analysis.
We use the density measurement to understand how system performance varies when compared against references using different extractive strategies by subdividing \DATASET into three subsets by extractiveness and evaluating using ROUGE on each.
We show here a similar analysis using the remaining two measures, coverage and compression.
Results for subsets based on coverage and compression are shown in Tables~\ref{table:coveragesubsets} and~\ref{table:compressionsubsets}.

\TABLESUBSETSCOVERAGE
\TABLESUBSETSCOMPRESSION

%% file: newsroom.bbl
\begin{thebibliography}{}
\expandafter\ifx\csname natexlab\endcsname\relax\def\natexlab#1{#1}\fi

\bibitem[{Bahdanau et~al.(2014)Bahdanau, Cho, and Bengio}]{Bahdanau2014}
Dzmitry Bahdanau, Kyunghyun Cho, and Yoshua Bengio. 2014.
\newblock \href{http://arxiv.org/abs/1409.0473}{Neural machine translation by
  jointly learning to align and translate}.
\newblock {\em CoRR\/} abs/1409.0473.
\newblock \url{http://arxiv.org/abs/1409.0473}.

\bibitem[{Barrios et~al.(2016)Barrios, L{\'{o}}pez, Argerich, and
  Wachenchauzer}]{Barrios2016}
Federico Barrios, Federico L{\'{o}}pez, Luis Argerich, and Rosa Wachenchauzer.
  2016.
\newblock \href{http://arxiv.org/abs/1602.03606}{Variations of the similarity
  function of textrank for automated summarization}.
\newblock {\em CoRR\/} abs/1602.03606.
\newblock \url{http://arxiv.org/abs/1602.03606}.

\bibitem[{Chen et~al.(2016)Chen, Bolton, and Manning}]{Chen2016}
Danqi Chen, Jason Bolton, and Christopher~D. Manning. 2016.
\newblock \href{http://aclweb.org/anthology/P/P16/P16-1223.pdf}{A thorough
  examination of the cnn/daily mail reading comprehension task}.
\newblock In {\em Proceedings of the 54th Annual Meeting of the Association for
  Computational Linguistics, {ACL} 2016, August 7-12, 2016, Berlin, Germany,
  Volume 1: Long Papers\/}.
\newblock \url{http://aclweb.org/anthology/P/P16/P16-1223.pdf}.

\bibitem[{Cheng and Lapata(2016)}]{Cheng2016}
Jianpeng Cheng and Mirella Lapata. 2016.
\newblock \href{http://aclweb.org/anthology/P/P16/P16-1046.pdf}{Neural
  summarization by extracting sentences and words}.
\newblock In {\em Proceedings of the 54th Annual Meeting of the Association for
  Computational Linguistics, {ACL} 2016, August 7-12, 2016, Berlin, Germany,
  Volume 1: Long Papers\/}.
\newblock \url{http://aclweb.org/anthology/P/P16/P16-1046.pdf}.

\bibitem[{Cho et~al.(2014)Cho, van Merrienboer, Gulcehre, Bahdanau, Bougares,
  Schwenk, and Bengio}]{Cho2014}
Kyunghyun Cho, Bart van Merrienboer, Caglar Gulcehre, Dzmitry Bahdanau, Fethi
  Bougares, Holger Schwenk, and Yoshua Bengio. 2014.
\newblock \href{https://doi.org/10.3115/v1/D14-1179}{Learning phrase
  representations using {RNN} encoder--decoder for statistical machine
  translation}.
\newblock In {\em Proceedings of the 2014 Conference on Empirical Methods in
  Natural Language Processing (EMNLP)\/}.
\newblock \url{https://doi.org/10.3115/v1/D14-1179}.

\bibitem[{Chopra et~al.(2016)Chopra, Auli, and Rush}]{Chopra2016}
Sumit Chopra, Michael Auli, and Alexander~M. Rush. 2016.
\newblock \href{http://aclweb.org/anthology/N/N16/N16-1012.pdf}{Abstractive
  sentence summarization with attentive recurrent neural networks}.
\newblock In {\em {NAACL} {HLT} 2016, The 2016 Conference of the North American
  Chapter of the Association for Computational Linguistics: Human Language
  Technologies, San Diego California, USA, June 12-17, 2016\/}. pages 93--98.
\newblock \url{http://aclweb.org/anthology/N/N16/N16-1012.pdf}.

\bibitem[{Dang(2006)}]{Dang2006}
Hoa~Trang Dang. 2006.
\newblock \href{http://dl.acm.org/citation.cfm?id=1654679.1654689}{Duc 2005:
  Evaluation of question-focused summarization systems}.
\newblock In {\em Proceedings of the Workshop on Task-Focused Summarization and
  Question Answering\/}. Association for Computational Linguistics,
  Stroudsburg, PA, USA, SumQA '06, pages 48--55.
\newblock \url{http://dl.acm.org/citation.cfm?id=1654679.1654689}.

\bibitem[{Dorr et~al.(2003)Dorr, Zajic, and Schwartz}]{Dorr2003}
Bonnie Dorr, David Zajic, and Richard Schwartz. 2003.
\newblock \href{https://doi.org/10.3115/1119467.1119468}{Hedge trimmer: A
  parse-and-trim approach to headline generation}.
\newblock In {\em Proceedings of the HLT-NAACL 03 on Text Summarization
  Workshop - Volume 5\/}. Association for Computational Linguistics,
  Stroudsburg, PA, USA, HLT-NAACL-DUC '03, pages 1--8.
\newblock \url{https://doi.org/10.3115/1119467.1119468}.

\bibitem[{Durrett et~al.(2016)Durrett, Berg-Kirkpatrick, and
  Klein}]{Durrett2016}
Greg Durrett, Taylor Berg-Kirkpatrick, and Dan Klein. 2016.
\newblock \href{http://www.aclweb.org/anthology/P16-1188}{Learning-based
  single-document summarization with compression and anaphoricity constraints.}
\newblock The Association for Computer Linguistics.
\newblock \url{http://www.aclweb.org/anthology/P16-1188}.

\bibitem[{Filippova et~al.(2015)Filippova, Alfonseca, Colmenares, Kaiser, and
  Vinyals}]{Filippova2015}
Katja Filippova, Enrique Alfonseca, Carlos~A. Colmenares, Lukasz Kaiser, and
  Oriol Vinyals. 2015.
\newblock \href{http://aclweb.org/anthology/D/D15/D15-1042.pdf}{Sentence
  compression by deletion with lstms.}
\newblock In Lluís Màrquez, Chris Callison-Burch, Jian Su, Daniele Pighin,
  and Yuval Marton, editors, {\em EMNLP\/}. The Association for Computational
  Linguistics, pages 360--368.
\newblock \url{http://aclweb.org/anthology/D/D15/D15-1042.pdf}.

\bibitem[{Filippova and Altun(2013)}]{Filippova2013}
Katja Filippova and Yasemin Altun. 2013.
\newblock \href{http://aclweb.org/anthology/D/D13/D13-1155.pdf}{Overcoming the
  lack of parallel data in sentence compression}.
\newblock In {\em Proceedings of the 2013 Conference on Empirical Methods in
  Natural Language Processing, {EMNLP} 2013, 18-21 October 2013, Grand Hyatt
  Seattle, Seattle, Washington, USA, {A} meeting of SIGDAT, a Special Interest
  Group of the {ACL}\/}. pages 1481--1491.
\newblock \url{http://aclweb.org/anthology/D/D13/D13-1155.pdf}.

\bibitem[{G{\"{u}}l{\c{c}}ehre et~al.(2016)G{\"{u}}l{\c{c}}ehre, Ahn,
  Nallapati, Zhou, and Bengio}]{Glehre2016}
{\c{C}}aglar G{\"{u}}l{\c{c}}ehre, Sungjin Ahn, Ramesh Nallapati, Bowen Zhou,
  and Yoshua Bengio. 2016.
\newblock \href{http://aclweb.org/anthology/P/P16/P16-1014.pdf}{Pointing the
  unknown words} \url{http://aclweb.org/anthology/P/P16/P16-1014.pdf}.

\bibitem[{Harman and Over(2004)}]{Harman2004}
Donna Harman and Paul Over. 2004.
\newblock \href{http://www.aclweb.org/anthology/W04-1003}{The effects of human
  variation in duc summarization evaluation}.
\newblock \url{http://www.aclweb.org/anthology/W04-1003}.

\bibitem[{Hermann et~al.(2015)Hermann, Ko\v{c}isk\'{y}, Grefenstette, Espeholt,
  Kay, Suleyman, and Blunsom}]{Hermann2015}
Karl~Moritz Hermann, Tom\'{a}\v{s} Ko\v{c}isk\'{y}, Edward Grefenstette, Lasse
  Espeholt, Will Kay, Mustafa Suleyman, and Phil Blunsom. 2015.
\newblock \href{http://dl.acm.org/citation.cfm?id=2969239.2969428}{Teaching
  machines to read and comprehend}.
\newblock In {\em Proceedings of the 28th International Conference on Neural
  Information Processing Systems - Volume 1\/}. MIT Press, Cambridge, MA, USA,
  NIPS'15, pages 1693--1701.
\newblock \url{http://dl.acm.org/citation.cfm?id=2969239.2969428}.

\bibitem[{Hong and Nenkova(2014)}]{Hong2014}
Kai Hong and Ani Nenkova. 2014.
\newblock \href{http://aclweb.org/anthology/E/E14/E14-1075.pdf}{Improving the
  estimation of word importance for news multi-document summarization}.
\newblock In {\em Proceedings of the 14th Conference of the European Chapter of
  the Association for Computational Linguistics, {EACL} 2014, April 26-30,
  2014, Gothenburg, Sweden\/}. pages 712--721.
\newblock \url{http://aclweb.org/anthology/E/E14/E14-1075.pdf}.

\bibitem[{Lin(2004{\natexlab{a}})}]{Lin2004b}
C.~Y. Lin. 2004{\natexlab{a}}.
\newblock Looking for a few good metrics: Automatic summarization evaluation -
  how many samples are enough?
\newblock In {\em Proceedings of the NTCIR Workshop 4\/}.

\bibitem[{Lin(2004{\natexlab{b}})}]{Lin2004}
Chin-Yew Lin. 2004{\natexlab{b}}.
\newblock
  \href{http://research.microsoft.com/~cyl/download/papers/WAS2004.pdf}{Rouge:
  A package for automatic evaluation of summaries}.
\newblock In {\em Proc. ACL workshop on Text Summarization Branches Out\/}.
  page~10.
\newblock \url{http://research.microsoft.com/~cyl/download/papers/WAS2004.pdf}.

\bibitem[{Manning et~al.(2014)Manning, Surdeanu, Bauer, Finkel, Bethard, and
  McClosky}]{Manning2014}
Christopher~D. Manning, Mihai Surdeanu, John Bauer, Jenny Finkel, Steven~J.
  Bethard, and David McClosky. 2014.
\newblock \href{http://www.aclweb.org/anthology/P/P14/P14-5010}{The {Stanford}
  {CoreNLP} natural language processing toolkit}.
\newblock In {\em Association for Computational Linguistics (ACL) System
  Demonstrations\/}. pages 55--60.
\newblock \url{http://www.aclweb.org/anthology/P/P14/P14-5010}.

\bibitem[{Mihalcea and Tarau(2004)}]{Mihalcea04}
R.~Mihalcea and P.~Tarau. 2004.
\newblock \href{http://www.aclweb.org/anthology/W04-3252}{{TextRank}: Bringing
  order into texts}.
\newblock In {\em Proceedings of {EMNLP-04}and the 2004 Conference on Empirical
  Methods in Natural Language Processing\/}.
\newblock \url{http://www.aclweb.org/anthology/W04-3252}.

\bibitem[{Nallapati et~al.(2017)Nallapati, Zhai, and Zhou}]{Nallapati2017}
Ramesh Nallapati, Feifei Zhai, and Bowen Zhou. 2017.
\newblock
  \href{http://aaai.org/ocs/index.php/AAAI/AAAI17/paper/view/14636}{Summarunner:
  {A} recurrent neural network based sequence model for extractive
  summarization of documents}.
\newblock In {\em Proceedings of the Thirty-First {AAAI} Conference on
  Artificial Intelligence, February 4-9, 2017, San Francisco, California,
  {USA.}\/}. pages 3075--3081.
\newblock \url{http://aaai.org/ocs/index.php/AAAI/AAAI17/paper/view/14636}.

\bibitem[{Nallapati et~al.(2016{\natexlab{a}})Nallapati, Zhou, dos Santos,
  G{\"{u}}l{\c{c}}ehre, and Xiang}]{Nallapati2016b}
Ramesh Nallapati, Bowen Zhou, C{\'{\i}}cero~Nogueira dos Santos, {\c{C}}aglar
  G{\"{u}}l{\c{c}}ehre, and Bing Xiang. 2016{\natexlab{a}}.
\newblock \href{http://aclweb.org/anthology/K/K16/K16-1028.pdf}{Abstractive
  text summarization using sequence-to-sequence rnns and beyond}.
\newblock In {\em Proceedings of the 20th {SIGNLL} Conference on Computational
  Natural Language Learning, CoNLL 2016, Berlin, Germany, August 11-12,
  2016\/}. pages 280--290.
\newblock \url{http://aclweb.org/anthology/K/K16/K16-1028.pdf}.

\bibitem[{Nallapati et~al.(2016{\natexlab{b}})Nallapati, Zhou, and
  Ma}]{Nallapati2016}
Ramesh Nallapati, Bowen Zhou, and Mingbo Ma. 2016{\natexlab{b}}.
\newblock \href{http://arxiv.org/abs/1611.04244}{Classify or select: Neural
  architectures for extractive document summarization}.
\newblock {\em CoRR\/} abs/1611.04244.
\newblock \url{http://arxiv.org/abs/1611.04244}.

\bibitem[{Napoles et~al.(2012)Napoles, Gormley, and Van~Durme}]{Napoles2012}
Courtney Napoles, Matthew Gormley, and Benjamin Van~Durme. 2012.
\newblock \href{http://dl.acm.org/citation.cfm?id=2391200.2391218}{Annotated
  gigaword}.
\newblock In {\em Proceedings of the Joint Workshop on Automatic Knowledge Base
  Construction and Web-scale Knowledge Extraction\/}. Association for
  Computational Linguistics, Stroudsburg, PA, USA, AKBC-WEKEX '12, pages
  95--100.
\newblock \url{http://dl.acm.org/citation.cfm?id=2391200.2391218}.

\bibitem[{Page et~al.(1999)Page, Brin, Motwani, and Winograd}]{Page1999}
Lawrence Page, Sergey Brin, Rajeev Motwani, and Terry Winograd. 1999.
\newblock \href{http://ilpubs.stanford.edu:8090/422/}{The pagerank citation
  ranking: Bringing order to the web.}
\newblock Technical Report 1999-66, Stanford InfoLab.
\newblock \url{http://ilpubs.stanford.edu:8090/422/}.

\bibitem[{Paulus et~al.(2017)Paulus, Xiong, and Socher}]{Paulus2017}
Romain Paulus, Caiming Xiong, and Richard Socher. 2017.
\newblock \href{http://arxiv.org/abs/1705.04304}{A deep reinforced model for
  abstractive summarization}.
\newblock {\em CoRR\/} abs/1705.04304.
\newblock \url{http://arxiv.org/abs/1705.04304}.

\bibitem[{Peters(2015)}]{Peters2015}
Matt Peters. 2015.
\newblock
  \href{https://moz.com/devblog/benchmarking-python-content-extraction-/algorithms-dragnet-readability-goose-and-eatiht/}{Benchmarking
  python content extraction algorithms: Dragnet, readability, goose, and
  eatiht}.
\newblock
  \url{https://moz.com/devblog/benchmarking-python-content-extraction-/algorithms-dragnet-readability-goose-and-eatiht/}.

\bibitem[{Rush et~al.(2015)Rush, Chopra, and Weston}]{Rush2015}
Alexander~M. Rush, Sumit Chopra, and Jason Weston. 2015.
\newblock \href{http://aclweb.org/anthology/D/D15/D15-1044.pdf}{A neural
  attention model for abstractive sentence summarization}.
\newblock In {\em Proceedings of the 2015 Conference on Empirical Methods in
  Natural Language Processing, {EMNLP} 2015, Lisbon, Portugal, September 17-21,
  2015\/}. pages 379--389.
\newblock \url{http://aclweb.org/anthology/D/D15/D15-1044.pdf}.

\bibitem[{Sandhaus(2008)}]{NYTimesCorpus2008}
E.~Sandhaus. 2008.
\newblock {The New York Times Annotated Corpus}.
\newblock {\em Linguistic Data Consortium, Philadelphia\/} 6(12).

\bibitem[{See et~al.(2017)See, Liu, and Manning}]{See2017}
Abigail See, Peter~J. Liu, and Christopher~D. Manning. 2017.
\newblock \href{https://doi.org/10.18653/v1/P17-1099}{Get to the point:
  Summarization with pointer-generator networks}.
\newblock In {\em Proceedings of the 55th Annual Meeting of the Association for
  Computational Linguistics, {ACL} 2017, Vancouver, Canada, July 30 - August 4,
  Volume 1: Long Papers\/}. pages 1073--1083.
\newblock \url{https://doi.org/10.18653/v1/P17-1099}.

\bibitem[{Sutskever et~al.(2014)Sutskever, Vinyals, and Le}]{Sutskever2014}
Ilya Sutskever, Oriol Vinyals, and Quoc~V. Le. 2014.
\newblock Sequence to sequence learning with neural networks.
\newblock In {\em Neural Information Processing Systems\/}.

\bibitem[{Tan et~al.(2017)Tan, Wan, and Xiao}]{Tan2017}
Jiwei Tan, Xiaojun Wan, and Jianguo Xiao. 2017.
\newblock \href{http://www.aclweb.org/anthology/P17-1108}{Abstractive document
  summarization with a graph-based attentional neural model}.
\newblock In {\em ACL\/}.
\newblock \url{http://www.aclweb.org/anthology/P17-1108}.

\bibitem[{Tiedemann(2012)}]{Tiedemann2012}
J{\"o}rg Tiedemann. 2012.
\newblock Parallel data, tools and interfaces in opus.
\newblock In {\em LREC\/}. volume 2012, pages 2214--2218.

\bibitem[{Tu et~al.(2016)Tu, Lu, Liu, Liu, and Li}]{Tu2016}
Zhaopeng Tu, Zhengdong Lu, Yang Liu, Xiaohua Liu, and Hang Li. 2016.
\newblock \href{https://doi.org/10.18653/v1/P16-1008}{Modeling coverage for
  neural machine translation}.
\newblock In {\em Proceedings of the 54th Annual Meeting of the Association for
  Computational Linguistics (Volume 1: Long Papers)\/}. Association for
  Computational Linguistics, pages 76--85.
\newblock \url{https://doi.org/10.18653/v1/P16-1008}.

\bibitem[{Vinyals et~al.(2015)Vinyals, Fortunato, and Jaitly}]{Vinyals2015}
Oriol Vinyals, Meire Fortunato, and Navdeep Jaitly. 2015.
\newblock \href{http://papers.nips.cc/paper/5866-pointer-networks.pdf}{Pointer
  networks}.
\newblock In C.~Cortes, N.~D. Lawrence, D.~D. Lee, M.~Sugiyama, and R.~Garnett,
  editors, {\em Advances in Neural Information Processing Systems 28\/}. Curran
  Associates, Inc., pages 2692--2700.
\newblock \url{http://papers.nips.cc/paper/5866-pointer-networks.pdf}.

\end{thebibliography}
